\documentclass[10pt,twocolumn,letterpaper]{article}

\usepackage{iccv}
\usepackage{times}
\usepackage{epsfig}
\usepackage{graphicx}
\usepackage{amsmath}
\usepackage{amssymb}
\usepackage{caption}
\usepackage{xcolor}
\usepackage{booktabs}

\usepackage{enumitem}
\usepackage{cite}

\usepackage[pagebackref=true,breaklinks=true,letterpaper=true,colorlinks,bookmarks=false]{hyperref}

\iccvfinalcopy %

\def\iccvPaperID{4548} %

\ificcvfinal\fi

\newif\ifdraft
\drafttrue

\newcommand{\Method}{DreamBooth3D}
\newcommand{\method}{DB3D}

\begin{document}

\title{\Method: Subject-Driven Text-to-3D Generation}

\author{Amit Raj \quad
Srinivas Kaza \quad
Ben Poole \quad
Michael Niemeyer \quad
Nataniel Ruiz\quad
Ben Mildenhall\quad \\
Shiran Zada\quad
Kfir Aberman\quad
Michael Rubinstein \quad
Jonathan Barron \quad
Yuanzhen Li\quad
Varun Jampani\quad \\
\textbf{Google}
}

\twocolumn[{%
\renewcommand\twocolumn[1][]{#1}%
\vspace{-2em}
\maketitle
\thispagestyle{empty}
\vspace{-2.5em}
\begin{center}
    \centering
    \includegraphics[width=\linewidth]{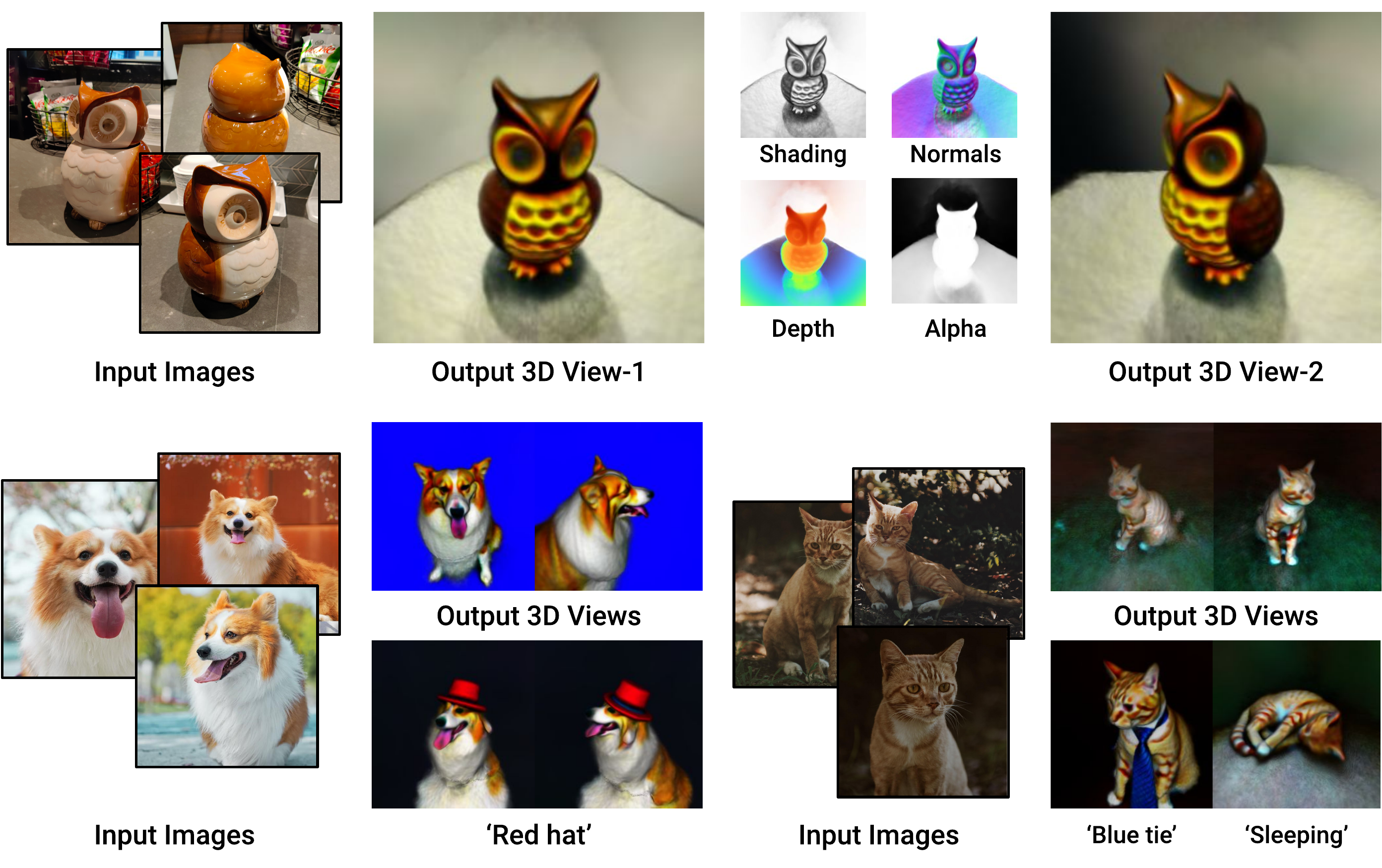}%
    \captionof{figure}
    {
    {\bf\Method} is a personalized text-to-3D generative model that creates plausible 3D assets of a specific subject from just 3-6 images. Top: 3D output and geometry estimated for an owl object. Bottom: our approach can generate variations of the 3D subject in different contexts (sleeping) or with different accessories (hat or tie) based on a text prompt. 
    }
    \label{fig:teaser}
\end{center}%
}]
\ificcvfinal\thispagestyle{empty}\fi

\begin{abstract}
\vspace{-4mm}
We present \Method, an approach to personalize text-to-3D generative models from as few as 3-6 casually captured images of a subject. Our approach combines recent advances in personalizing text-to-image models (DreamBooth) with text-to-3D generation (DreamFusion). We find that na\"ively combining these methods fails to yield satisfactory subject-specific 3D assets due to personalized text-to-image models overfitting to the input viewpoints of the subject.
We overcome this through a 3-stage optimization strategy where we jointly leverage the 3D consistency of neural radiance fields together with the personalization capability of text-to-image models.
Our method can produce high-quality, subject-specific 3D assets with text-driven modifications such as novel poses, colors and attributes that are not seen in any of the input images of the subject. More results are available at our project page:
\url{https://dreambooth3d.github.io}
\end{abstract}
\vspace{-2mm}
\section{Introduction}
\vspace{-2mm}

Text-to-Image (T2I) generative models~\cite{imagen,dalle2,ldm,muse} have greatly expanded the ways we can create and edit visual content. 
Recent works~\cite{poole2022dreamfusion,magic3d,latent-nerf,jacobiannet} have demonstrated high-quality Text-to-3D generation by optimizing neural radiance fields (NeRFs)~\cite{nerf} using the T2I diffusion models. Such automatic 3D asset creation with input text prompts alone has applications in a wide range of areas, such as graphics, VR, movies, and gaming.

Although text prompts allow for some degree of control over the generated 3D asset, it is often difficult to precisely control its identity, geometry, and appearance solely with text. 
In particular, these methods lack the ability to generate 3D assets of a specific subject (e.g., a specific dog instead of a generic dog). Enabling the generation of subject-specific 3D assets would significantly ease the workflow for artists and 3D acquisition.
There has been remarkable success~\cite{dreambooth_cite,TI,customdiffusion} in personalizing T2I models for subject-specific 2D image generation. These techniques allow the generation of specific subject images in varying contexts, but they do not generate 3D assets or afford any 3D control, such as viewpoint changes.

In this work, we propose `\Method', a method for subject-driven Text-to-3D generation. Given a few (3-6) casual image captures of a subject (without any additional information such as camera pose), we generate subject-specific 3D assets that also adhere to the contextualization provided in the input text prompts. That is, we can generate 3D assets with geometric and appearance identity of a given subject while also respecting the variations (e.g. sleeping or jumping dog) provided by the input text prompt.

For \Method, we draw inspiration from the recent works~\cite{poole2022dreamfusion} which propose optimizing a NeRF model using a loss derived from T2I diffusion models.
We observe that simply personalizing a T2I model for a given subject and then using that model to optimize a NeRF is prone to several failure modes.
A key issue is that the personalized T2I models tend to overfit to the camera viewpoints that are only present in the sparse subject images. As a result, the resulting loss from such personalized T2I models is not sufficient to optimize a coherent 3D NeRF asset from arbitrary continuous viewpoints.

With \Method, we propose an effective optimization scheme where we optimize both a NeRF asset and T2I model in conjunction with each other to jointly make them subject-specific.
We leverage DreamFusion~\cite{poole2022dreamfusion} for NeRF optimization and use DreamBooth~\cite{dreambooth_cite} for T2I model fine-tuning.
Specifically, we propose a 3-stage optimization framework where in the first stage, we partially finetune a DreamBooth model and then use DreamFusion to optimize a NeRF asset.
The partially finetuned DreamBooth model does not overfit to the given subject views, but also do not capture all the subject-specific details. So the resulting NeRF asset is 3D coherent, but is not subject-specific.
In the second stage, we fully finetune a DreamBooth model to capture fine subject details and use that model to create multiview pseudo-subject images. That is, we translate multiview renderings from the trained NeRF into subject images using the fully-trained DreamBooth model.
In the final stage, we further optimize the DreamBooth model using both the given subject images along with the pseudo multi-view images; which is then used to optimize our final NeRF 3D volume.
In addition, we also use a weak reconstruction loss over the pseudo multi-view dataset to further regularize the final NeRF optimization.
The synergistic optimization of the NeRF and T2I models prevents degenerate solutions and avoids overfitting of the DreamBooth model to specific views of the subject, while ensuring that the resulting NeRF model is faithful to the subject's identity.

For experimental analysis, we use the dataset of 30 subjects proposed in DreamBooth~\cite{dreambooth_cite} which uses the same input setting of sparse casual subject captures.
Results indicate our approach can generate realistic 3D assets with high likeness to a given subject while also respecting the contexts present in the input text prompts. Fig.~\ref{fig:teaser} shows sample results of \Method~ on different subjects and contextualizations.
When compared to several baselines, both quantitative and qualitative results demonstrate that \Method~ generations are more 3D coherent and better capture subject details.

\section{Related Works}

\noindent \textbf{Text-to-Image Generation.}
Earlier works on generative models are dominated by Generative Adversarial Networks (GANs) which train a generator to synthesis images that are indistinguishable from real images ~\cite{goodfellow2020generative, sauer2023stylegan}. Other generative approaches include autoregressive models that generate images pixel by pixel or patch by patch~\cite{yu2022scaling,gafni2022make} and
masked image models that iteratively predict the marginal distribution of masked patches in the image~\cite{muse,chang2022maskgit}. Recently, denoising diffusion models ~\cite{ho2020denoising} have been proposed for image synthesis, which can generate high-quality images by iteratively denoising a noise image toward a clean image \cite{imagen,dalle2,ldm,dhariwal2021diffusion}. Diffusion models can also be conditioned on various inputs such as depth-map~\cite{control_net}, sketch~\cite{voynov2022sketch}, semantic segmentation~\cite{ldm, bar2023multidiffusion}, text\cite{nichol2022glide,imagen,dalle2,ldm} and others \cite{huang2023composer, control_net, li2023gligen}.
For text conditioning, these models take advantage of pre-trained large language models (LLMs)~\cite{clip,t5_transformer} in order to generate images that are aligned with a natural language text prompt given by the user.
Motivated by the success of T2I diffusion models, many works utilize pre-trained T2I models for various tasks such as text-based image manipulation \cite{kawar2022imagic,mokady2022null,brooks2022instructpix2pix}.

\vspace{1mm}
\noindent \textbf{3D Generation.}
First works on learning-based 3D content generation performed 3D reconstruction from one or multiple images~\cite{Choy2016ECCV,Mescheder2019CVPR,Fan2017CVPR,Wen2019ICCV,Gkioxari2019ICCV}.
While leading to good reconstruction results, they require large-scale datasets of accurate 3D data for training which limits their use in real-world scenarios.
Another line of work~\cite{Schwarz2021NEURIPS,piGAN2021,Niemeyer2021CVPR,egthreed,Gu2022ICLR} circumvents the need for accurate 3D data by training 3D-aware generative models from image collections.
While achieving impressive results, these methods are sensitive to the assumed pose distribution and restricted to single object classes. 
Very recently, text-to-3D methods~\cite{dreamfields,poole2022dreamfusion,magic3d,latent-nerf} have been proposed that can generate 3D assets from text prompts by utilizing large pretrained T2I diffusion models.
In many applications, however, the conditioning are rather input images optionally with text instead of pure text.
As a result, multiple works investigate how input images can be incorporated into the optimization pipeline, \eg by applying a reconstruction loss on the input image and predicted monocular depth~\cite{Xu2022ARXIV,Deng2022ARXIV} or a predicted object mask~\cite{real_fusion}.
This, however, limits their use as it does not exploit the full strength of diffusion models, \eg, the object cannot be recontextualized with additional text input.   
Instead, we propose to not directly reconstruct the input image, but rather the concept of the provided object. This allows not only for reconstruction, but also for recontextualization and more, and the input images do not need to be taken with the same background, lighting, camera \etc.

\vspace{1mm}
\noindent \textbf{Subject-driven Generation.}
Recent advances in subject-driven image generation \cite{dreambooth_cite, TI, customdiffusion} enable users to personalize their image generation for specific subjects and concepts. This has provided T2I models with the ability to capture the visual essence of specific subjects and synthesize novel renditions of them in different contexts. DreamBooth \cite{dreambooth_cite} accomplishes this by expanding the language-vision dictionary of the model using rare tokens, model finetuning, and a prior preservation loss for regularization. Textual Inversion \cite{TI} accomplishes this by optimizing for a new "word" in the embedding space of a pre-trained text-to-image model that represents the input concept. It's worth noting that these methods do not generate 3D assets or 3D coherent images. There have also been developments in guiding image generation with grounding inputs \cite{li2023gligen}, editing instructions \cite{brooks2022instructpix2pix}, and task-specific conditions such as edges, depth, and surface normals \cite{control_net}. However, these techniques do not provide personalization to specific subjects, and do not generate 3D assets.

 \section{Approach}
\vspace{-2mm}

\newcommand{\XNoiseEq}[1]{\mathcal{D}_{\theta}(\alpha_t {#1} + \sigma_t \mathbf{\epsilon}, \mathbf{c}) - {#1}}
\newcommand{\DiffNet}[0]{\hat{\mathcal{D}}_\theta}
\newcommand{\pseudo}[0]{\gamma}

\noindent \textbf{Problem setup.}

The input to our approach forms a set of $k$ casual subject captures, each with $n$ pixels, $\{{I_i \in \mathbb{R}^{n \times 3}}\}$ ($i \in \{1,...,k\}$) and a text prompt $T$ for the contextualization or semantic variation (e.g., sleeping vs. standing dog). Our aim is to generate a 3D asset that captures the identity (geometry and appearance) of the given subject while also being faithful to the text prompt. We optimize 3D assets in the form of Neural Radiance Fields (NeRF)~\cite{nerf}, which consists of an MLP network $\mathcal{M}$ that encodes radiance fields in a 3D volume. Note that this problem is considerably more under-constrained and challenging compared to a typical 3D reconstruction setting that requires multi-view image captures. We build our technique on recent advances in T2I personalization and Text-to-3D optimization. Specifically, we use DreamFusion~\cite{poole2022dreamfusion} text-to-3D optimization and DreamBooth~\cite{dreambooth_cite} personalization in our framework, which we briefly review next.

\subsection{Preliminaries}

\noindent \textbf{DreamBooth T2I Personalization.}
T2I diffusion models such as Imagen~\cite{imagen}, StableDiffusion~\cite{ldm} and DALL-E 2~\cite{dalle2} generate images from any given text prompt. In particular, a T2I diffusion model $ \mathcal{D}_{\theta}(\mathbf{\epsilon}, \mathbf{c})$ takes as input an initial noise $\epsilon \sim \mathcal{N}(0,1)$ and a text embedding $\mathbf{c}=\Theta(T)$ for a given prompt $T$ with a text encoder $\Theta$ and generates an image that follows the description of the prompt. The images generated from these T2I models are usually consistent with the prompt, however, it is difficult to exert fine-grained control in the generated images. To that end, DreamBooth~\cite{dreambooth_cite} proposes a simple yet effective approach to personalize a T2I diffusion model by finetuning the network on a small set of casual captures $\{I_i\}$. 

Briefly, DreamBooth uses the following diffusion loss function to finetune the T2I model:
 \begin{equation}
     \mathcal{L}_{d} = \mathbb{E}_{\epsilon, t}\!\left[ w_t \left\| \XNoiseEq{I_i} \right\|^2 \right] \, ,
     \label{eqn:diff}
 \end{equation}

where $t \sim \mathcal{U}[0,1]$ denotes the time-step in the diffusion process and $w_t$, $\alpha_t$ and $\sigma_t$ are the corresponding scheduling parameters.
Optionally, DreamBooth uses the class prior-preserving loss for improved diversity and to avoid language drift. Refer to~\cite{dreambooth_cite} for additional details.

\begin{figure*}[h!]
    \centering
    \includegraphics[width=17.5cm]{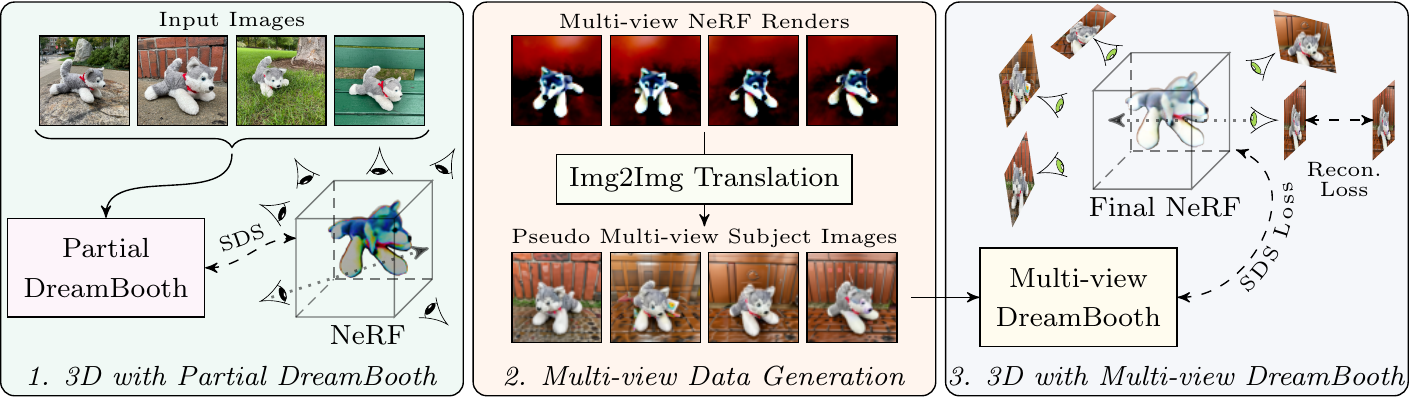}
    \caption{\textbf{DreamBooth3D Overview}. In the stage-1 (left), we first partially train a DreamBooth and use the resulting model to optimize the initial NeRF. In stage-2 (middle), we render multi-view images along random viewpoints from the initial NeRF and then translate them into pseudo multi-view subject images using a fully-trained DreamBooth model. In the final stage-3 (right), we further fine-tune the partial DreamBooth using multi-view images and then use the resulting multi-view DreamBooth to optimize the final NeRF 3D asset using the SDS loss along with the multi-view reconstruction loss.}
    \label{fig:approach}
\end{figure*}

\vspace{1mm}
\noindent \textbf{DreamFusion} optimizes a volume represented as a NeRF $\mathcal{M}_\phi$ with parameters $\phi$ so that random views of the volume match a text prompt $T$ using a T2I diffusion model. The learned implicit network $\mathcal{M}_\phi$ maps from a 3D location to an albedo and density. The normals computed from the gradient of the density are used to randomly relight the model to improve geometric realism with Lambertian shading. Given a random view $v$, and random lighting direction, we perform volume rendering to output a shaded image $\hat{I}_v$. To optimize the parameters of the NeRF $\phi$ so that these images look like a text prompt $T$, DreamFusion introduced score distillation sampling (SDS) that pushes noisy versions of the rendered images to lower energy states of the T2I diffusion model:

\begin{equation}
    \nabla_\phi \mathcal{L}_\mathit{SDS} = \mathbb{E}_{\epsilon, t} \left[w_t\left( \XNoiseEq{\hat{I}_v} \right) \frac{\partial \hat{I}_v}{\partial \phi}\right].
    \label{eqn:sds}
\end{equation}
By randomizing over views and backpropagating through the NeRF, it encourages the renderings to look like an image produced by T2I model $\mathcal{D}_\theta$ for a given text prompt. DreamFusion proposes to use coarse view-based prompting to optimize NeRF along multiple views. We follow the exact settings used in~\cite{poole2022dreamfusion} for all experiments.

\subsection{Failure of Naive Dreambooth+Fusion}
\label{sec:dreambooth_fusion}
A straight-forward approach for subject-driven text-to-3D generation is first personalizing a T2I model and then use the resulting model for Text-to-3D optimization. For instance, doing DreamBooth optimization followed by DreamFusion. which we refer to as DreamBooth+Fusion.
Similar baselines are also explored with preliminary experiments in some very recent works such as~\cite{latent-nerf,magic3d}.
However, we find that naive DreamBooth+Fusion technique results in unsatisfactory results as shown in Fig.~\ref{fig:visual_comparisons}.
A key issue we find is that DreamBooth tends to overfit to the subject views that are present in the training views, leading to reduced viewpoint diversity in the image generations.
Subject likeness increases with more DreamBooth finetuning steps, while the generated viewpoints get close to that of input exemplar views.
As a result, the SDS loss on such a DreamBooth model is not sufficient to obtain a coherent 3D NeRF asset. In general, we observe that the DreamBooth+Fusion NeRF models have
same subject views (e.g., face of a dog) imprinted across different viewpoints, a failure mode denoted the ``Janus problem''~\cite{poole2022dreamfusion}.

\subsection{Dreambooth3D Optimization}
\label{sec:optim}

To mitigate the aforementioned issues, we propose an effective multi-stage optimization scheme called \Method~for subject-driven text-to-3D generation. Fig.~\ref{fig:approach} illustrates the 3 stages in our approach, which we describe in detail next.

\vspace{1mm}
\noindent \textbf{Stage-1: 3D with Partial DreamBooth.}
We first train a personalized DreamBooth model $\hat{\mathcal{D}}_\theta$ on the input subject images such as those shown in Fig.~\ref{fig:approach} (left).
Our key observation is that the initial checkpoints of DreamBooth (partially finetuned) T2I models do not overfit to the given subject views. DreamFusion on such partially finetuned DreamBooth models can produce a more coherent 3D NeRF.
Specifically, we refer to the partially trained DreamBooth model as $\hat{\mathcal{D}}_\theta^{partial}$ and use the SDS loss (Eq.~\ref{eqn:sds}) to optimize an initial NeRF asset for a given text prompt as illustrated in Fig.~\ref{fig:approach} (left).
However, the partial DreamBooth model as well as the NeRF asset lack complete likeness to the input subject.
We can see this \textit{initial} NeRF output in stage-1 to be a 3D model of the subject class that has partial likeness to the given subject while also being faithful to the given text prompt.

\vspace{1mm}
\noindent \textbf{Stage-2: Multi-view Data Generation.}
This stage forms an important part of our approach, where we make use of 3D consistent initial NeRF together with the fully-trained DreamBooth to generate pseudo multi-view subject images.
Specifically, we first render multiple images $\{\hat{I}_v\ \in \mathbb{R}^{n \times 3}\}$ along random viewpoints $\{v\}$ from the initial NeRF asset resulting in the multi-view renders as shown in Fig.~\ref{fig:approach} (middle).
We then add a fixed amount of noise by running the forward diffusion process from each render to $t_\mathit{pseudo}$, and then run the reverse diffusion process to generate samples using the fully-trained DreamBooth model $\hat{\mathcal{D}}_\theta$ as in \cite{SDEdit}. This sampling process is run independently for each view, and results in images that represent the subject well, and cover a wide range of views due to the conditioning on the  noisy render of our initial NeRF asset. However, these images are not multi-view consistent as the reverse diffusion process can add different details to different views, so we call this collection of images {\em pseudo} multi-view images.

Fig.~\ref{fig:approach}~(middle) shows sample resulting images from this image to image (Img2Img) translation. Some prior works such as~\cite{SDEdit} use such Img2Img translations for image editing applications. In contrast, we use the Img2Img translation in combination with DreamBooth and NeRF 3D asset to generate pseudo multi-view subject images.
A key insight in this stage is that DreamBooth can effectively generate unseen views of the subject given that initial images are close to those unseen views. In addition, DreamBooth can effectively generate output images with more likeness to the given subject compared to input noisy images.
Fig.~\ref{fig:approach}~(middle) shows sample outputs of Img2Img translation with the DreamBooth demonstrating more likeness to the subject images while also preserving the viewpoints of the input NeRF renders.

\vspace{1mm}
\noindent \textbf{Stage-3: Final NeRF with Multi-view DreamBooth.}
The previous stage provides pseudo multi-view subject images $\{I_v^{\mathit{pseudo}}\}$ with \textit{near-accurate} camera viewpoints $\{v\}$.
Both the viewpoints as well as the subject-likeness are only approximately accurate due to the stochastic nature of DreamBooth and Img2Img translation.
We combine the generated multi-view images $\{I_v^{\mathit{pseudo}}\}$ along with the input subject images $\{I_i\}$ to create a combined data $\mathcal{I}^{aug}=\{{I}_v^{\mathit{pseudo}}\} \cup \{I_i\}$.
We then use this data to optimize our final DreamBooth model followed by a final NeRF 3D asset.

More concretely, we further finetune the partially trained DreamBooth $\hat{\mathcal{D}}_\theta^*$ from stage-1 using this augmented data resulting in a DreamBooth we refer to as Multi-view DreamBooth $\hat{\mathcal{D}}_\theta^\mathit{multi}$.
We then use this $\hat{\mathcal{D}}_\theta^\mathit{multi}$ model to optimize NeRF 3D asset using the DreamFusion SDS loss (Eq.\ref{eqn:sds}).
This results in a NeRF model with considerably better subject-identity as the multi-view DreamBooth has better view generalization and subject preservation compared to the partial DreamBooth from stage-1.

In practice, we observe that the resulting NeRF asset, optimized only using SDS loss, usually has good geometry-likeness to the given subject but has some color saturation artifacts.
To account for the color shift we introduce a novel weak reconstruction loss using our psuedo multi-view images $\{{I}_v^{\mathit{pseudo}}\}$. In particular, since we know the camera parameters $\{P_v\}$ from which these images were generated, we additionally regularize the training of the second NeRF MLP $\mathcal{F}_\gamma$, with $\gamma$ parameters with the reconstruction loss:
\begin{equation}
 \mathcal{L}_\mathit{recon} = \left\| \Gamma(\mathcal{F}_\gamma, P_v) - I_v^{\mathit{pseudo}} \right\|_p,
\label{eqn:recon}
\end{equation}
Where $\Gamma(\mathcal{F}_\gamma, P_v)$ is the rendering function that renders an image from the NeRF $\mathcal{F}_\gamma$ along the camera viewpoint $P_v$.
This loss serves the dual purpose of pulling the color distribution of the generated volume closer to those of the image exemplars and to improve subject likeness in unseen views.
Fig.~\ref{fig:approach} (right) illustrate the optimization of final NeRF with SDS and multi-view reconstruction losses. 
The final NeRF optimization objective is given as:
\begin{equation}
 \mathcal{L} =  \lambda_\mathit{recon} \mathcal{L}_\mathit{recon} + \lambda_\mathit{SDS} \mathcal{L}_\mathit{SDS} + \lambda_\mathit{nerf} \mathcal{L}_\mathit{nerf},
\end{equation}
where $\mathcal{L}_\mathit{nerf}$ denotes the additional NeRF regularizations used in Mip-NeRF360~\cite{Barron2021MipNeRF3U}. 
See the supplementary material for additional details of the DreamBooth3D optimization.
\section{Experiments}

\begin{figure*}[h!]
    \centering
    \includegraphics[width=\linewidth]{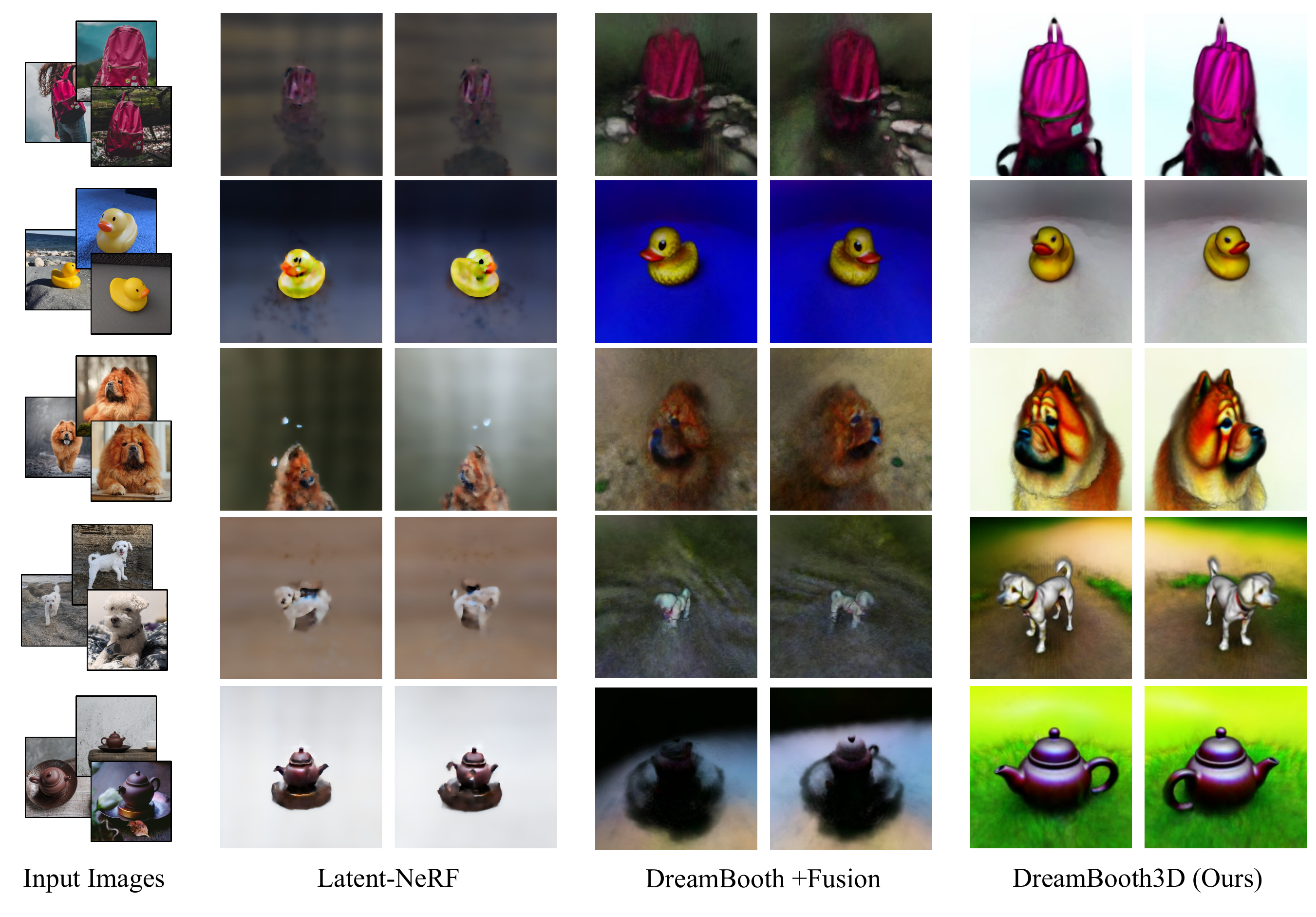}
    \caption{\textbf{Visual Results} on 5 different subjects with two baseline techniques of Latent-NeRF and DreamBooth+Fusion along with those of our technique (DreamBooth3D).
    Results clearly indicate better 3D consistent results with our approach compared to either of the baseline techniques. See the supplement for additional visualizations and videos.}
    \label{fig:visual_comparisons}
\end{figure*}

\begin{figure*}[h!]
    \centering
    \includegraphics[width=\linewidth]{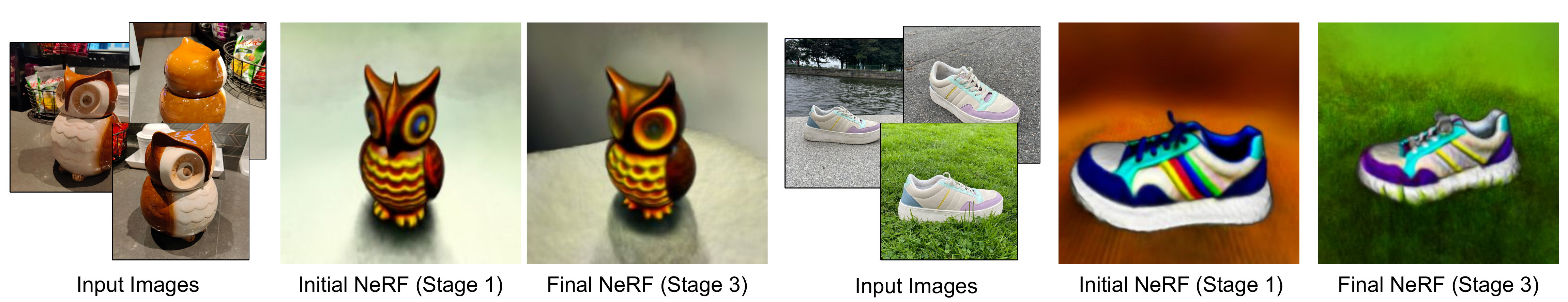}
    \caption{\textbf{Initial vs. Final NeRF Estimates}. Sample multi-view results show that the initial NeRF obtained after stage-1 has only a partial likeness to the given subject whereas the final NeRF from stage-3 of our pipeline has better subject-identity.}
    \label{fig:init_final_nerf}
\end{figure*}

\noindent \textbf{Implementation Details}.
We use the Imagen~\cite{imagen} 
T2I model in our experiments. The Imagen model uses the T5-XXL\cite{t5_transformer} language model for text encoding.
On the NeRF side, we use DreamFusion~\cite{poole2022dreamfusion}. Our model takes around 3 hours per prompt to complete all the 3 stages of the optimization on a 4 core TPUv4. 
We use a fixed 150 iterations to train the partial DreamBooth model $\hat{\mathcal{D}}_{\theta}^{partial}$.
For the full DreamBooth $\hat{\mathcal{D}}_\theta$ training, we use 800 iterations,
which we find to be optimal across different subjects.  
We render 20 images uniformly sampled at a fixed radius from the origin for pseudo multi-view data generation. We finetune the partially trained $\hat{\mathcal{D}}_\theta^*$ for additional 150 iterations in Stage 3. Refer to the supplementary material for more hyperparameter details.

\vspace{1mm}
\noindent \textbf{Datasets}.
We train our personalized text to 3D models on the image collections released by the authors of \cite{dreambooth_cite}. This dataset consists of 30 different image collections with 4-6 casual captures of a wide variety of subjects (dogs, toys, backpack, sunglasses, cartoon etc.). 
We additionally capture few images of some rare objects (like ``owl showpiece'' in Fig.~\ref{fig:init_final_nerf}) to analyze performance on rare objects. Further, we optimize each 3D model on 3--6 prompts to demonstrate 3D contextualizations.

\vspace{1mm}
\noindent \textbf{Baselines}.
We consider two main baselines for comparisons. 
Latent-NeRF \cite{latent-nerf} which learns a 3D NeRF model on a latent feature space instead of in RGB pixel space, using an SDS loss in the latent space of Stable Diffusion~\cite{ldm}.
As a baseline, we run Latent-NeRF using the fully dreamboothed T2I model and refer to it as ``Latent-NeRF'' or ``L-NeRF" in our experiments.
We further compare against a single stage DreamFusion+DreamBooth approach where we first train a DreamBooth diffusion model followed by 3D NeRF optimization using DreamFusion.
We refer to our results as ``\Method'' or ``\method'' in the experiments.

\vspace{1mm}
\noindent \textbf{Evaluation Metrics}. 
We evaluate our approach with the CLIP R-Precision metric, which measures how accurately we can retrieve a text prompt from an image \cite{park2021benchmark}. Similar to \cite{poole2022dreamfusion}, we compute the average CLIP R-Precision over 160 evenly spaces azimuth renders at a fixed elevation of 40 degrees. The CLIP models used for evaluation are the CLIP ViT-B/16, ViT-B/32, and ViT-L-14 models.
Since these CLIP metrics can only approximately capture the quality and subject-fidelity of the generated 3D assets, we additionally perform user studies comparing different results.

\subsection{Results}
\vspace{-2mm}

\noindent \textbf{Visual Results}. 
Fig.~\ref{fig:teaser} shows sample visual results of our approach along with different semantic variations and contextualizations. Results demonstrate high-quality geometry estimation with \Method~for even our uncommon owl object.
Contexualization examples demonstrate that \Method~faithfully respects the context present in the input text prompt. Fig.~\ref{fig:visual_comparisons} shows sample results of our approach in comparison to those of Latent-NeRF and DreamBooth+Fusion baselines. Even though Latent-NeRF works reasonably well in some cases (such as rubber duck in Fig.~\ref{fig:visual_comparisons}), more often it fails to converge to a coherent 3D model with reasonable shapes.
In several cases, DreamBooth+Fusion usually produces the 3D assets with Janus problem (same appearance and geometry imprinted across different view angles).
\Method, on other hand, consistently produces 360$^{\circ}$ consistent 3D assets while capturing both the geometric and appearance details of the given subject.

\vspace{1mm}
\noindent \textbf{Quantitative Comparisons}.
Table.~\ref{tab:r_precision_metrics} shows CLIP R-precision metrics for naive DreamBooth+Fusion (as baseline) and our DreamBooth3D generations. Results clearly demonstrate significantly higher scores for the DreamBooth3D results indicating better 3D consistency and text-prompt alignment of our results.

\vspace{1mm}
\noindent \textbf{Initial vs. Final NeRF}.
Fig.~\ref{fig:init_final_nerf} shows sample initial and final NeRF results generated after stages 1 and 3 of our pipeline. As the visual results illustrate, initial-NeRFs only have partial likeness to the given subject, but are consistent in 3D. The final NeRFs from the stage-3 has better likeness to the given subject while retaining the consistent 3D structure. These examples demonstrate the need for the 3-stage optimization in \Method.

\begin{figure}[h!]
    \centering
    \includegraphics[width=\columnwidth]{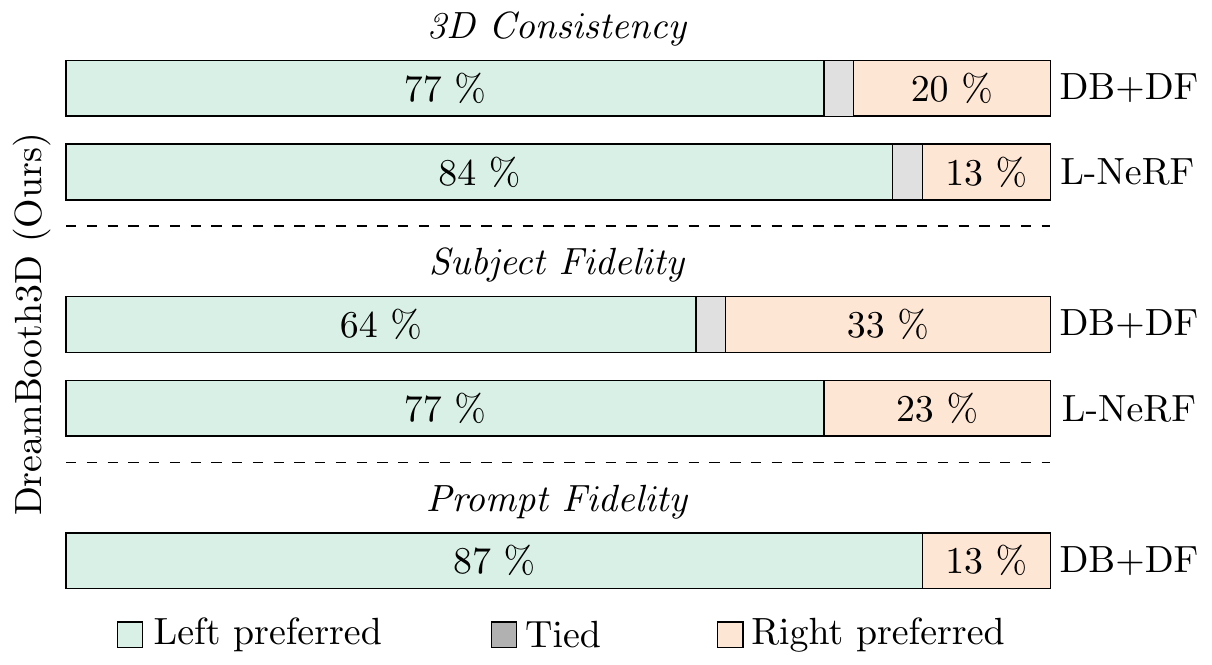}
    \caption{\textbf{User Study}. Users show a significant preference for our DreamBooth3D over DB+DF and L-NeRF for 3D consistency, subject fidelity and prompt fidelity.}
    \label{fig:user_study}
\end{figure}

\vspace{1mm}
\noindent \textbf{User Study}. We conduct pairwise user studies comparing DreamBooth3D to baselines in order to evaluate our method under three axes: (1) Subject fidelity, where users are asked to answer the question ``Which 3D item looks more like the original subject?’’; (2) 3D consistency and plausibility where users answer ``Which 3D item has a more plausible and consistent geometry?’’ and (3) Prompt fidelity to the input prompts where users answer ``Which video best respects the provided prompt?". Users can choose either our method or the baseline, or a third option ``Cannot determine / both equally’’.
For the first two user studies on 3D consistency and subject fidelity we compare rotating video results, one for each of the 30 subjects in the dataset and ask 11 users to vote for each pair. For the prompt fidelity study, we generate videos for 54 unique prompt and subject pairs and ask 21 users to respond.
We compute final results using majority voting and present them in Figure~\ref{fig:user_study}.  We find that DreamBooth3D is significantly preferred over the baselines in terms of 3D consistency, subject fidelity as well as prompt fidelity.

\begin{figure}[h!]
    \centering
    \includegraphics[width=\columnwidth]{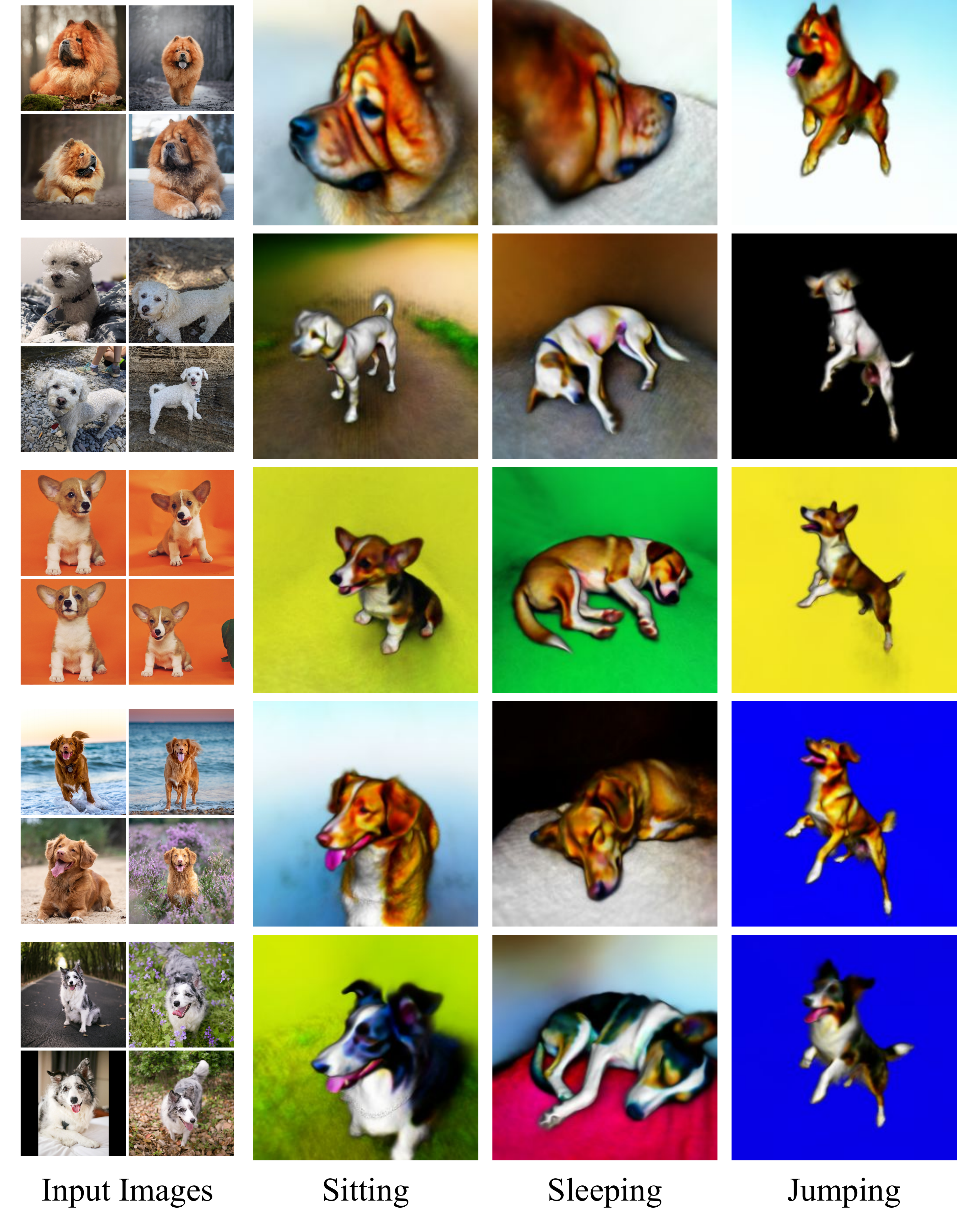}
    \caption{\textbf{3D Recontextualizations with DreamBooth3D}. With simple edits in the text prompt, we can generate non-rigid 3D articulations and deformations that correspond to the semantics of the input text. Visuals show consistent contexualization of different dogs in different contexts of sitting, sleeping and jumping. See the supplement for videos.}
    \label{fig:recontext}
\end{figure}

\vspace{1mm}

\subsection{Sample Applications}
\vspace{-2mm}

\Method~can faithfully represent the context present in the text prompts while also preserving the subject identity. With simple changes in the text-prompt, \Method~enables many interesting 3D applications, several of which would otherwise require tedious manual effort to tackle using traditional 3D modeling techniques.

\noindent \textbf{Recontextualization}.
Fig.~\ref{fig:recontext} shows sample results on different dog subjects, where we
recontextualize the 3D dog models with simple prompts of sitting, sleeping and jumping.
As the visuals demonstrate, the corresponding 3D models consistently respect the given context
in the text prompt across all the subjects. In addition, the 3D articulations and local deformations
in the output 3D models are highly realistic even though several of these poses are unseen in the
input subject images.

\begin{figure}[h!]
    \centering
    \includegraphics[width=\columnwidth]{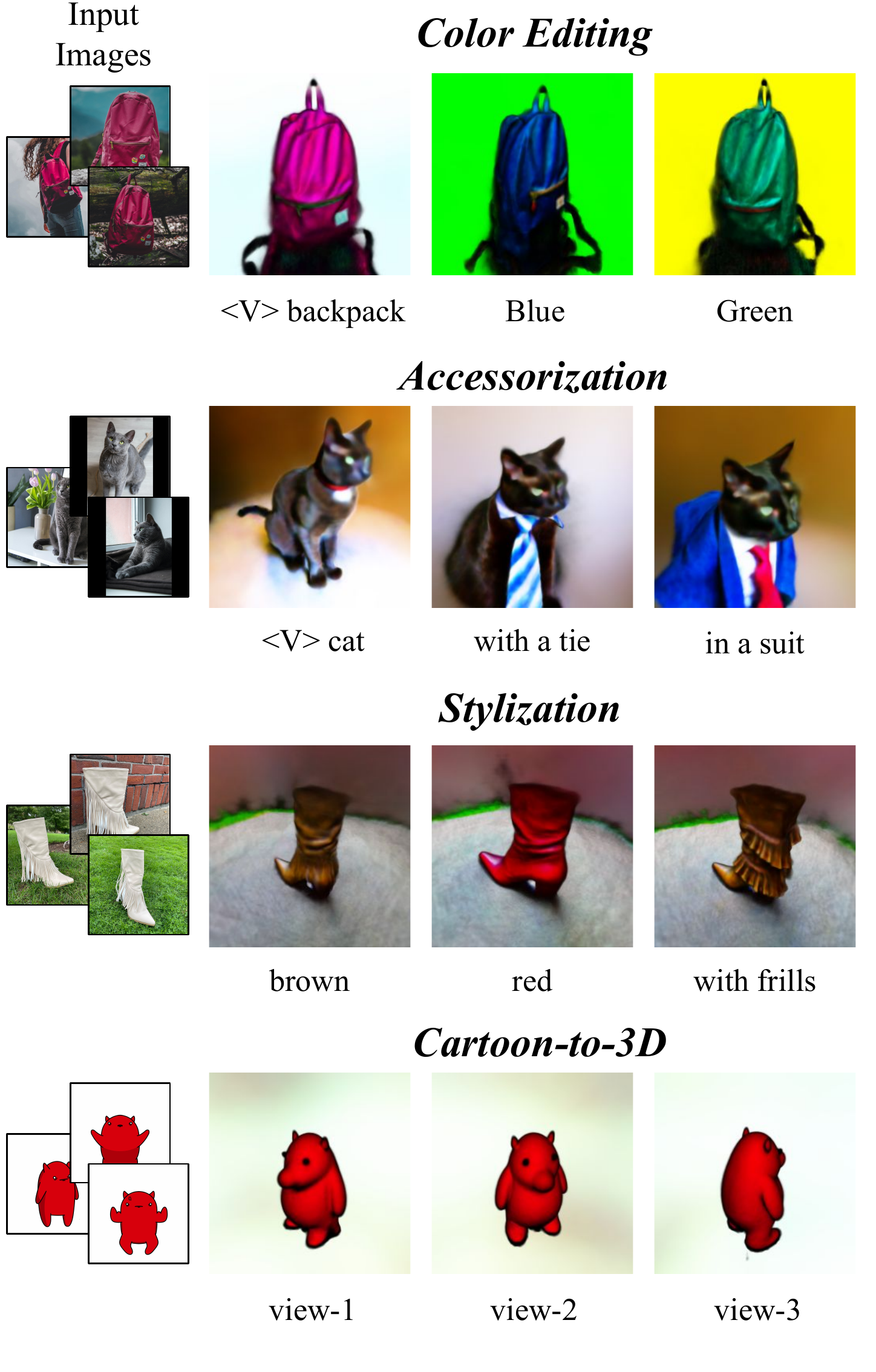}
    \caption{\textbf{Sample Applications}. DreamBooth3D's subject preservation and faithfulness to the text prompt enables several applications such as color/material editing, accessorization, stylization, \etc. \Method~can even produce plausible 3D models from unrealistic cartoon images. See the supplemental material for videos.}
    \label{fig:applications}
\end{figure}

\vspace{1mm}
\noindent \textbf{Color/Material Editing}.
Fig.~\ref{fig:applications} shows sample color editing results, where a pink backpack can be converted into a blue or green backpack with simple text prompts like `a [v] blue backpack'.
Similarly, one could also easily edit the material appearance of the 3D asset (for e.g., metal can to wodden can).
Refer to the supplementary material for more color and material editing results.

\vspace{1mm}
\noindent \textbf{Accessorization}.
Fig.~\ref{fig:applications} shows sample accessorization results on a cat subject, where we put on a tie or a suit into the 3D cat model output.
Likewise, one can think of other accessorizations like putting on a hat or sunglasses etc.

\vspace{1mm}
\noindent \textbf{Stylization}.
Fig.~\ref{fig:applications} also shows sample stylization results, where a cream colored shoe is stylized based on color and the addition of frills. 

\vspace{1mm}
\noindent \textbf{Cartoon-to-3D}.
A rather striking result we find during our experiments is that \Method~can even convert non-photorealistic subject images such as 2D flat cartoon images into plausible 3D shapes.
Fig.~\ref{fig:applications} shows a sample result where the resulting 3D model for the red cartoon character is plausible, even though all the images show the cartoon only from the front.
Refer to the supplementary material for more qualitative results on different applications.

\begin{table}[]
    \centering
    \small
    \begin{tabular}{@{\,\,}l@{\,\,\,\,}c@{\,\,\,\,}c@{\,\,\,\,}c@{\,\,}}
    \toprule
     & ViT-B/16$\uparrow$ & ViT-B/32$\uparrow$ & ViT-L-14$\uparrow$ \\
     \midrule
    DreamBooth+Fusion & 0.509 & 0.490 & 0.506 \\
    DreamBooth3D (Ours) & \textbf{0.783} & \textbf{0.710} & \textbf{0.797}  \\
    \bottomrule
    \end{tabular}
    \caption{\textbf{Quantitative comparisons} using CLIP R-precision on DreamBooth+Fusion (baseline) and DreamBooth3D generations indicate that renderings from our 3D model outputs more accurately resemble the text prompts.}
    \label{tab:r_precision_metrics}
\end{table}

\vspace{1mm}
\subsection{Limitations}
\begin{figure}[h!]
    \centering
    \includegraphics[width=\columnwidth]{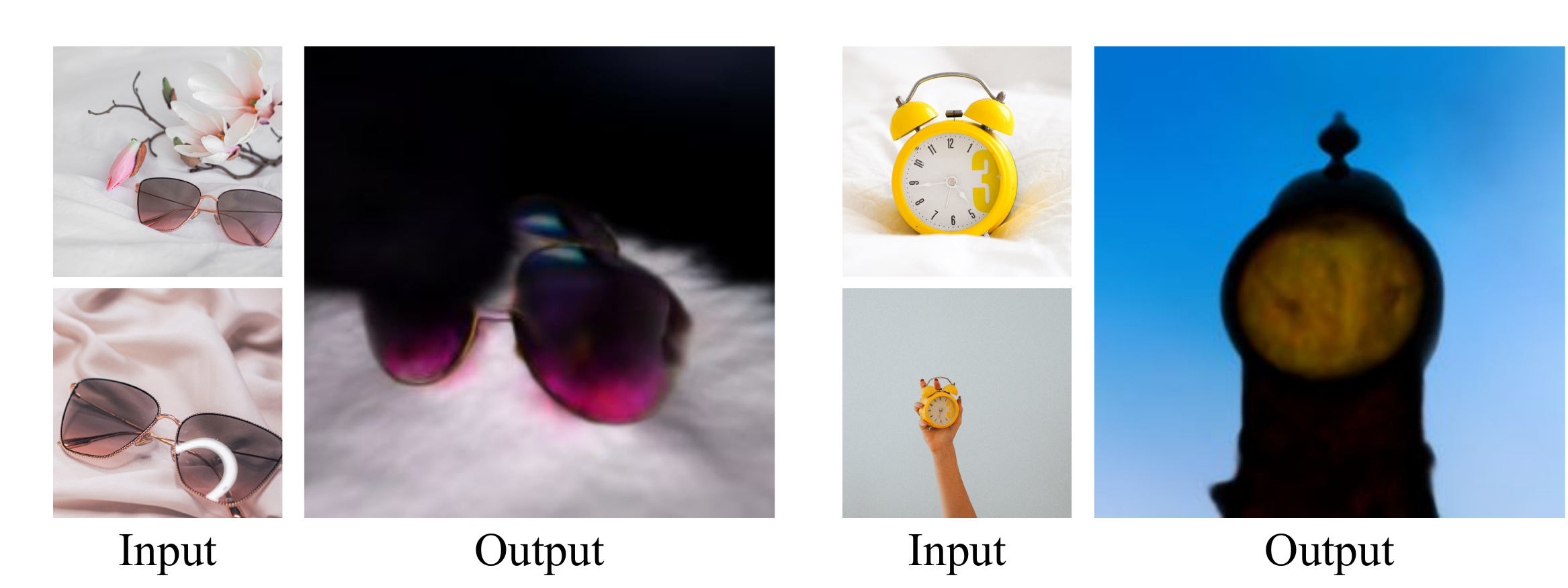}
    \caption{\textbf{Sample Failure Cases}. We observe~\Method~often fails to reconstruct thin object structures like sunglasses, and sometimes fails to reconstruct objects with not enough view variation in the input images.}
    \label{fig:limitations}
\end{figure}

While our method allows for high-quality 3D asset creation of a given subject and improves over prior work, we observe several limitations.
First, the optimized 3D representations are sometimes oversaturated and oversmoothed, which is partially caused by SDS-based optimization with high guidance weighting~\cite{poole2022dreamfusion}. This is also a result of being restricted to a relatively low image resolution of $64 \times 64$ pixels. Improvements in the efficiency of both diffusion models and neural rendering will potentially allow for scaling to higher resolutions.  
Furthermore, the optimized 3D representations can sometimes suffer from the Janus problem of appearing to be front-facing from multiple inconsistent viewpoints if the input images do not contain any viewpoint variations.
Finally, our model sometimes struggles to reconstruct thin object structures like sunglasses. 
Fig.~\ref{fig:limitations} shows a couple of failure results.

\section{Conclusion}
\vspace{-2mm}

In this paper, we have proposed \Method~, a method for subject-driven text-to-3D generation. Given a few (3-6) casual image captures of a subject (without any additional information such as camera pose), we generate subject-specific 3D assets that also adhere to the contextualization provided in the input text prompts (e.g. sleeping, jumping, red, etc.). 
Our extensive experiments on the DreamBooth dataset~\cite{dreambooth_cite} have shown that our method can generate realistic 3D assets with high likeness to a given subject while also respecting the contexts present in the input text prompts. Our method outperforms several baselines in both quantitative and qualitative evaluations. In the future, we plan to continue to improve the photorealism and controllability of subject-driven 3D generation.

{\small
\bibliographystyle{ieee_fullname}
\bibliography{egbib}
}
\appendix

\thispagestyle{empty}
\clearpage

\section{Summary Video with Visual Results}

We summarize our findings in a video, which outlines the three-stage DreamBooth3D method and includes a comparison to the baselines. We also show how our approach compares to other approaches via a user study. Finally, several example applications are shown, including material editing, accessorization, color changes, and pose changes.

\section{NeRF Details}
We use Mip-NeRF\cite{Barron2021MipNeRF3U} as our choice of volumetric representation. Particularly, to render the color of a ray $\mathbf{r}(t) = \mathbf{o} + t\mathbf{d}$ cast into the scene, Mip-NeRF divides the ray into intervals and for each interval calculates the mean and variance $(\mathbf{\mu},\Sigma)$ of a conical frustum corresponding to the interval. These values are then used to encode the ray using integrated  positional encoding 
\begin{equation}
    \gamma(\mu,\Sigma) = \left\{ \begin{bmatrix}
    \sin(2^l\mu)\exp(-2^{(2l-1)}\text{diag}(\sigma)) \\
    \sin(2^l\mu)\exp(-2^{(2l-1)}\text{diag}(\sigma))
    \end{bmatrix} \right\}_{l=0}^L
\end{equation}
The learnt volume $\mathcal{N}_{\phi}$ is then used to generate albedo $\mathbf{c}$ and opacity $\sigma$. 
\[
\mathbf{c},\sigma=\mathcal{N}_\phi(\gamma(\mu,\Sigma))
\]
The final color is then calculated using numerical quadrature as in \cite{nerf}. As in \cite{poole2022dreamfusion}, we define $\Sigma=\lambda^2_t I$, where, $\lambda_t$ is annealed from a high to low value, to gradually introduce higher frequency components during the optimization. The NeRF volume is regularized using the orientation loss introduced in Ref-NeRF \cite{verbin2022ref}, to encourage better geometry. Particularly,
\begin{equation}
    \mathcal{L}_{ori} = \sum_i \text{stop\_grad}(w_i)\max(0,\mathbf{n}_i.\mathbf{v})
\end{equation}
Where $\mathbf{n}_i$ is the normal direction at a point, $w_i$ are rendering weights as defined in \cite{nerf} and $\mathbf{v}$ is the lighting direction. An additional opacity loss is used to encourage foreground/ background separation
\begin{equation}
\mathcal{L}_{op} = \sqrt{(\sum_i w_i)^2 +0.01}
\end{equation}
The final NeRF regularization loss is then given by:
\begin{equation}
    \mathcal{L}_{nerf} =  \mathcal{L}_{op} + \mathcal{L}_{ori}
\end{equation}

\section{Additional results}
\begin{figure*}[h!]
    \centering
    \includegraphics[width=\textwidth]{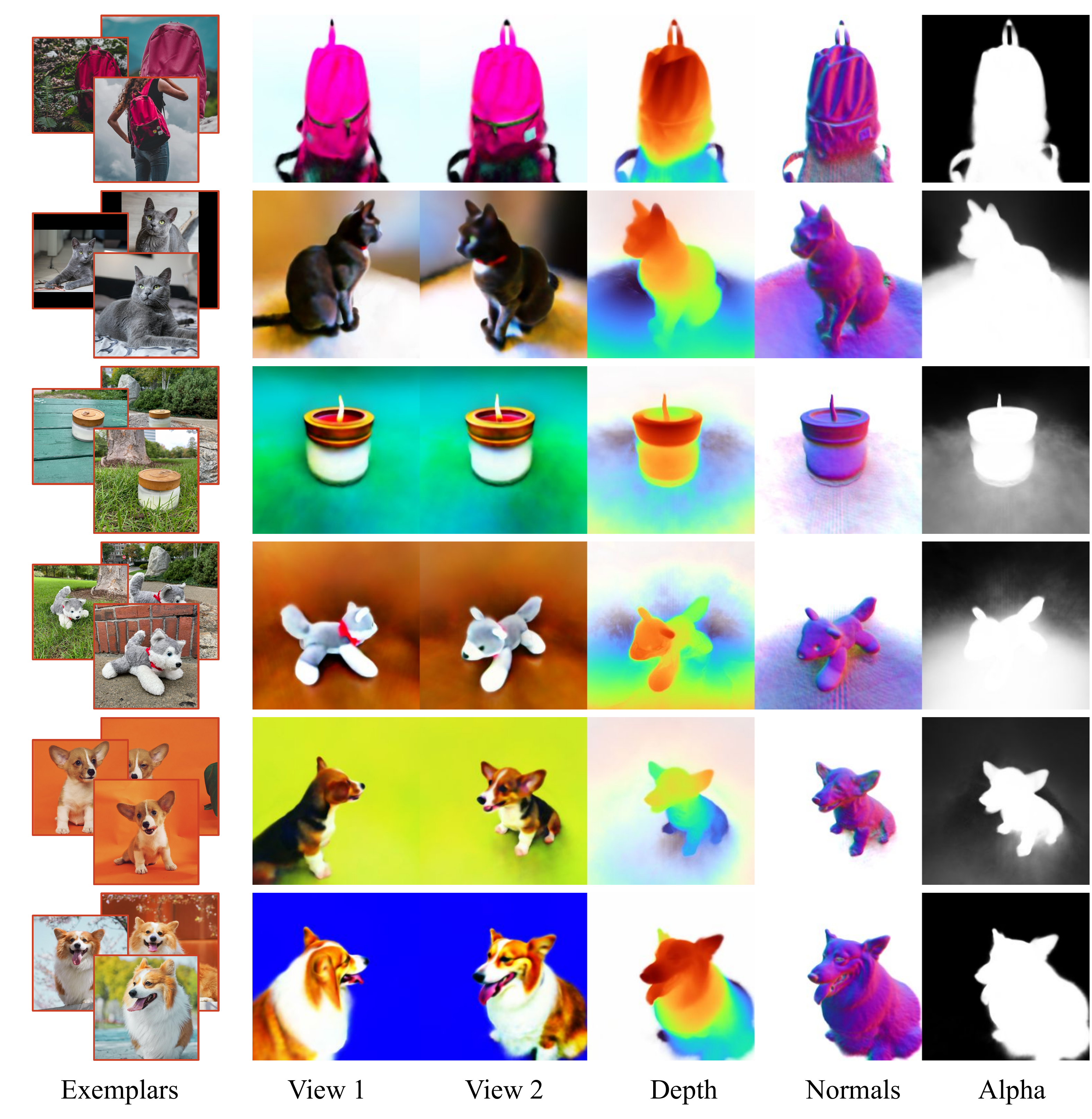}
    \caption{\textbf{Additional Results}. Dreambooth3D can produce 3D consistent volumes from text prompts. The figure shows generated assets for the base prompt "A photo of $<$v$>$" where $<$v$>$ is the subject presented in the first column. Column 2 and 3 shows two different views of the rendered volume. Column 3,4 and 5 shows the depth, normals and opacity of the second view respectively. }
    \label{fig:results}
    \vspace{3mm}
\end{figure*}

\begin{figure*}[h!]
    \centering
    \includegraphics[width=\textwidth]{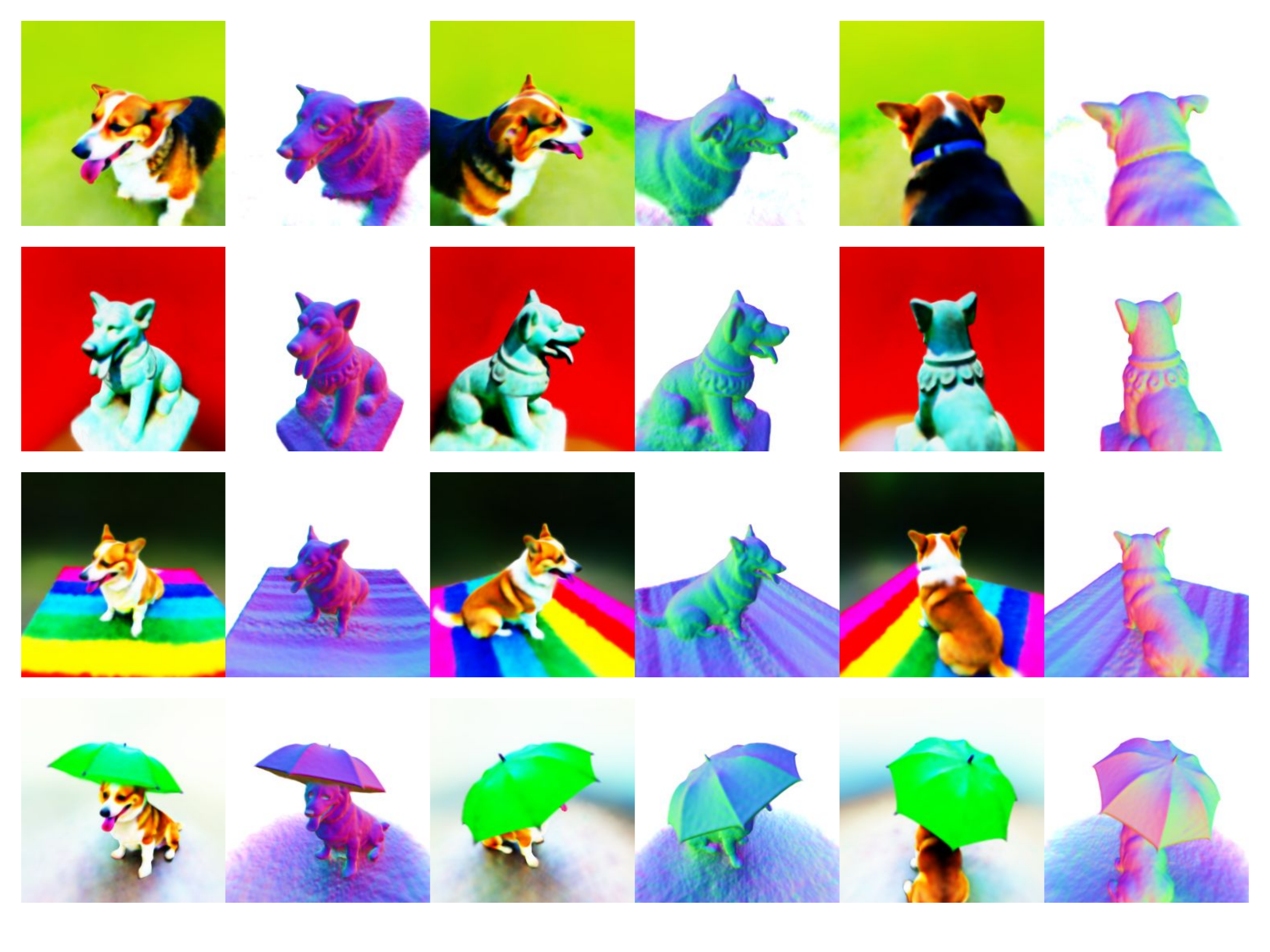}
    \caption{\textbf{Additional Results}. Dreambooth3D is capable of a number of accessorization, composition, material editing tasks through text prompting. An example of this form of prompting is "A photo of $<$v$>$ wearing a green umbrella". Row 1 shows the rendered geometry and normals of the base subject, and the subsequent rows show material-edited, composited, or accessorized variants. Row 2 demonstrates a material edit to change the dog into a stone statue. Row 3 composites a rainbow carpet into the scene. Row 4 adds a green umbrella.}
    \label{fig:results2}
\end{figure*}

Fig.~\ref{fig:results} provides additional results with associated depths, normals and alpha maps to demonstrate the 3D consistency of our results on a variety of subjects. Fig.~\ref{fig:results2} shows multiple views of the assets rendered for the same subject with different text prompts. 

\end{document}